\documentclass{article} 
\usepackage[T1]{fontenc}
\usepackage[utf8]{inputenc} 
\usepackage{CJKutf8}
\usepackage{iclr2019_conference,times}

\usepackage{amsmath,amsfonts,bm}

\def\eqref#1{equation~\ref{#1}}

\def\1{\bm{1}}

\DeclareMathAlphabet{\mathsfit}{\encodingdefault}{\sfdefault}{m}{sl}
\SetMathAlphabet{\mathsfit}{bold}{\encodingdefault}{\sfdefault}{bx}{n}

\usepackage{hyperref}
\usepackage{url}
\usepackage{wrapfig}
\usepackage{subcaption}
\usepackage{graphicx}
\usepackage{multirow}
\usepackage{listings}

\title{Adversarially-Trained Normalized Noisy-Feature Auto-Encoder for Text Generation}

\author{Xiang Zhang \\
Courant Institute of Mathematical Sciences, New York University \\
Element AI \\
\texttt{xiang@cs.nyu.edu}
\And
Yann LeCun \\
Courant Institute of Mathematical Sciences, New York University \\
Center for Data Science, New York University \\
Facebook AI Research, Facebook Inc. \\
\texttt{yann@cs.nyu.edu}
}

\begin{document}
\begin{CJK}{UTF8}{gbsn}

\maketitle

\begin{abstract}
This article proposes Adversarially-Trained Normalized Noisy-Feature Auto-Encoder (ATNNFAE) for byte-level text generation. An ATNNFAE consists of an auto-encoder where the internal code is normalized on the unit sphere and corrupted by additive noise. Simultaneously, a replica of the decoder (sharing the same parameters as the AE decoder) is used as the generator and fed with random latent vectors. An adversarial discriminator is trained to distinguish training samples reconstructed from the AE from samples produced through the random-input generator, making the entire generator-discriminator path differentiable for discrete data like text. The combined effect of noise injection in the code and shared weights between the decoder and the generator can prevent the mode collapsing phenomenon commonly observed in GANs. Since perplexity cannot be applied to non-sequential text generation, we propose a new evaluation method using the total variance distance between frequencies of hash-coded byte-level \(n\)-grams (NGTVD). NGTVD is a single benchmark that can characterize both the quality and the diversity of the generated texts. Experiments are offered in 6 large-scale datasets in Arabic, Chinese and English, with comparisons against \(n\)-gram baselines and recurrent neural networks (RNNs). Ablation study on both the noise level and the discriminator is performed. We find that RNNs have trouble competing with the \(n\)-gram baselines, and the ATNNFAE results are generally competitive.
\end{abstract}

\section{Introduction}

Learning high-level, abstract representations of text or other discrete structures is a task that may have many applications in NLP, including text generation, translation and general understanding. This article makes 4 contributions: (1) a new class of model and objective functions called Adversarially-Trained Normalized Noisy-Feature Auto-Encoder (ATNNFAE) that is suited for encoding and generating sequence of symbols, such as text; (2) a recursive convolutional architecture for the encoder and decoder/generator that is designed to represent texts of any length at the byte level; (3) a measure of performance for byte-level text generators called \(n\)-Gram Total Variation Distance (NGTVD) that compares statistics of hash-coded \(n\)-grams; (4) experimental results on text generation by training on very large text corpora in multiple languages.

\begin{figure}[t]
  \begin{center}
    \includegraphics[width=0.9\textwidth]{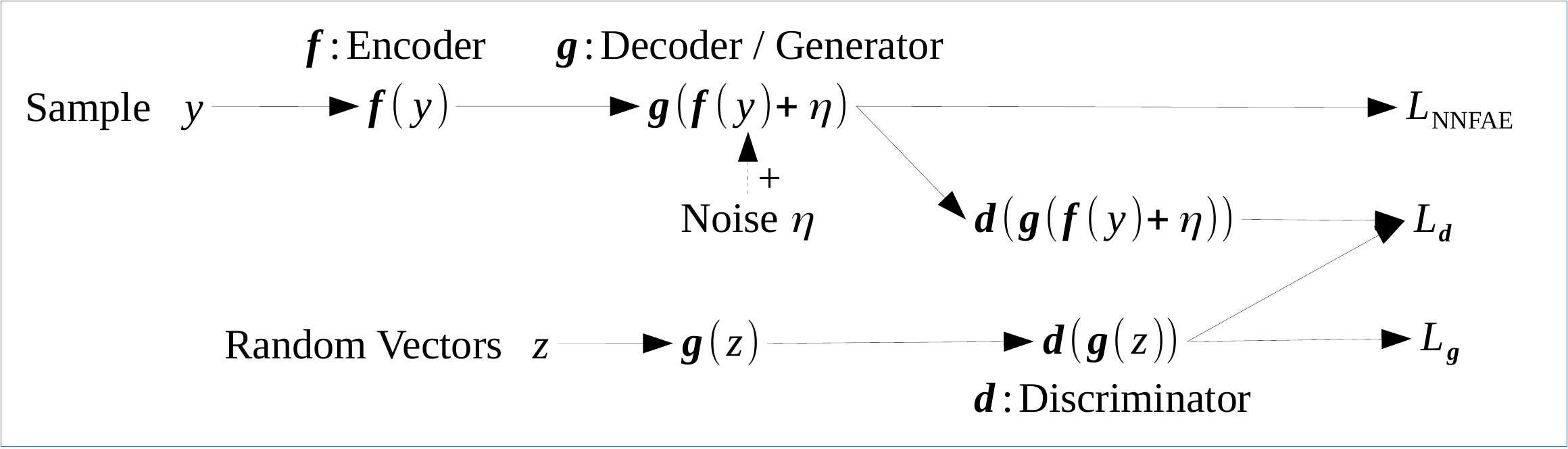}
  \end{center}
  \caption{Adversarially-Trained Normalized Noisy-Feature Auto-Encoder (ATNNFAE) combines Normalized Noisy-Feature Auto-Encoder (NNFAE) and GAN. Note that the NNFAE decoder and the GAN generator are the same model \(\boldsymbol g\). ATNNFAE learns by alternating between 3 objectives. (1) The NNFAE objective \(L_\textrm{NNFAE}\) optimizes the encoder \(\boldsymbol f\) and the decoder \(\boldsymbol g\) to reconstruct the sample \(y\) from the feature corrupted by the additive noise \(\eta\). (2) The discriminator objective \(L_{\boldsymbol d}\) optimizes the discriminator \(\boldsymbol d\) to distinguish between the reconstructed output \({\boldsymbol g}(f(y) + \eta)\) from the NNFAE and the generator output \({\boldsymbol g}(z)\), in which z is set of vectors uniformly sampled from the unit sphere. (3) The generator objective \(L_{\boldsymbol g}\) optimizes the generator \(\boldsymbol g\) to ``fool'' the discriminator by making \({\boldsymbol d}({\boldsymbol g}(z))\) approach the same target used for \({\boldsymbol d}({\boldsymbol g}({\boldsymbol f}(y) + \eta))\) in the discriminator loss \(L_{\boldsymbol d}\). }
  \label{fig:adve}
\end{figure}

The basic architecture of ATNNFAE, shown in figure \ref{fig:adve}, consists of an auto-encoder where the internal code is normalized on the unit sphere and corrupted by additive noise. The AE is trained to reconstruct the input while eliminating the effect of noise. This effectively regularizes the information content of the code and forces the AE to maximize the distance between the codes of training samples. Simultaneously, a replica of the decoder (sharing the same parameters as the AE decoder) is used as the generator and fed with random latent vectors, uniformly sampled on the unit sphere. An adversarial discriminator is trained to distinguish training samples reconstructed from the AE from samples produced through the random-input decoder replica, making the entire generator-discriminator path differentiable for discrete data like text. The combined effect of noise injection in the code and shared weights between the decoder and the generator can prevent the mode collapsing phenomenon commonly observed in GANs \citep{GPMXWOCB14}.

The auto-encoder architecture we used is a byte-level recursive convolutional auto-encoder \citep{ZL18}. This choice is made because convolutional networks have been shown to have better auto-encoding accruacy compared to recurrent neural networks (RNNs) at both word \citep{ZSWGHC17} and byte \citep{ZL18} levels. As a result of this choice, our model becomes a non-sequential (or non-autoregressive \citep{GBXLS18}) text generator. Since perplexity or bits-per-character cannot be directly applied to non-sequential text generation, we propose an evaluation method using the \(n\)-gram total variation distance (NGTVD). NGTVD can capture both the quality and the diversity of generated texts, since in either case it will result in a mismatch on the \(n\)-gram frequencies. Experiments are offered in 6 large scale datasets in Arabic, Chinese and English, with comparisons against \(n\)-gram baselines and recurrent neural networks (RNNs).

There are numerous attempts in text generation with or without GANs that merit discussion in this article. We discuss the difference between these ideas in section \ref{sec:rela}.  ATNNFAE is introduced in section \ref{sec:mode}. The NGTVD evaluation method is introduced in section \ref{sec:eval}. Section \ref{sec:expr} offers the experimental results, with comparisons against \(n\)-gram models and RNNs. Ablation study on the necessity of the discriminator and the denoising process is also included, which prompts us to do a hyper-parameter search on the level of noise. Furthermore, we showed additional improvements for RNNs and \(n\)-gram models via output selection, and for ATNNFAE models via \(n\)-gram correction. Before concluding this article, we also show some generated examples by interpolating in the feature space.

\section{Related Work}
\label{sec:rela}

The challenge of applying GAN to text lies in the gap between the discrete nature of text data and the continuous nature of the discriminator. Most solutions can be classified into 3 categories.

\begin{enumerate}
\item The discriminator accepts a discrete sample. Because it is not differentiable with respect to the generator, some other solutions are required to provide gradients to the generator.
\item The discriminator accepts some intermediate representation in the generator. It is differentiable with respect to the sub-network in the generator that produces this representation.
\item The discriminator accepts a continuous sample in some transformed space. Some network is required to transform a discrete sample to this space, but the entire path is differentiable.
\end{enumerate}

In the case that the discriminator accepts a discrete output, a few different approaches have been proposed. The idea of SeqGAN proposed by \citet{YZWY17} uses policy gradient \citep{SMSM00} to provide gradients to the generator, by casting the problem as a sequential decision making process. On the other hand, MaskGAN \citep{FGD18} uses a discriminator that accepts a discrete word with its surrounding context, using the same policy gradient method in an actor-critic framework \citep{SB98} \citep{DPS12}. Beyond reinforcement learning approaches, MaliGAN \citep{CLZHLSB17} uses the maximum likelihood principle by assuming the discriminator has achieved optimum with respect to the current generator.

There are numerous attempts to apply the discriminator to the some intermediate representation of the generator. Professor forcing \citep{GLZZCB16} was proposed to use GAN on the hidden units to ensure generator stability, which improves the quality of long samples. Adversarial feature matching \citep{ZGFCHSC17} was an idea to improve RNN generators using a convolutional discriminator on the hidden units. Adversarially regularized auto-encoder (ARAE) \citep{ZKZRL18} makes the generator match the feature from the encoder.

Our approach -- Adversarially-Trained Normalized Noisy-Feature Auto-Encoder (ATNNFAE) -- is one that belongs in the realm of letting the discriminator operate in some transformed sample space. Previously, \citet{KH16} proposed to use a Gumbel-softmax distribution on the output of an RNN while the samples are provided as one-hot vectors. This approach could collapse at large-scale, because the discriminator could easily distinguish between one-hot encoding and the generator's output. Instead, we use an auto-encoder to transform a one-hot encoded sample to an unnormalized log probability space.

Beyond using GANs, an alternative approach is to use the variational auto-encoder (VAE) framework \citep{KW13}. However, previous attempts such as \citet{BVVDJB16} have shown limited success. In VAE, the normalized feature from the encoder is optimized towards constant values, making it easy for the model to ignore the encoder. In ATTNFAE, the feature is corrupted with additive noise, and its strength is controllable via a hyper-parameter.

Similar to our approach, the generator in parallel WaveNet \citep{OLBSVKDLCS17} maps from a sequence of random vectors to samples. It has an implicit sequential dependence via inverse-autoregressive flows (IAF) \citep{KSJCSW16}. However, the parallel WaveNet paper \citep{OLBSVKDLCS17} only experimented on supervised tasks in speech synthesis, and it is unknown whether an unconditional generative model is possible.

Finally, none of the discussed approaches can prevent mode collapsing of GANs, while our method can do so via denoising in a normalized feature space. In addition, using GAN for non-sequential text generation is a necessity, which is in contrast with RNNs, for which the maximum likelihood principle (``teacher forcing'' \citep{WZ89}) already exists for training.

\section{Adversarially-Trained Normalized Noisy-Feature Auto-Encoder (ATNNFAE)}
\label{sec:mode}

This section introduces the different components in ATNNFAE, using byte-level recursive convolutional auto-encoders \citep{ZL18}. Additionally, the hyper-parameters used for training are detailed.

\subsection{Normalized Noisy-Feature Auto-Encoder (NNFAE)}

The NNFAE architecture in this article is the byte-level recursive convolutional auto-encoder \citep{ZL18}, chosen for its better accuracy compared to RNNs. Good auto-encoding accuracy is required because its output is used as the target to the discriminator. \citet{MLC18} offered improvements by removing the linear layers and using a fixed number of recursion groups, which give better results for long byte sequences. Following them, we use an NNFAE that has a fixed number of recursion groups without linear layers.

\begin{wrapfigure}{r}{0.5\textwidth}
  \begin{center}
    \includegraphics[width=0.5\textwidth]{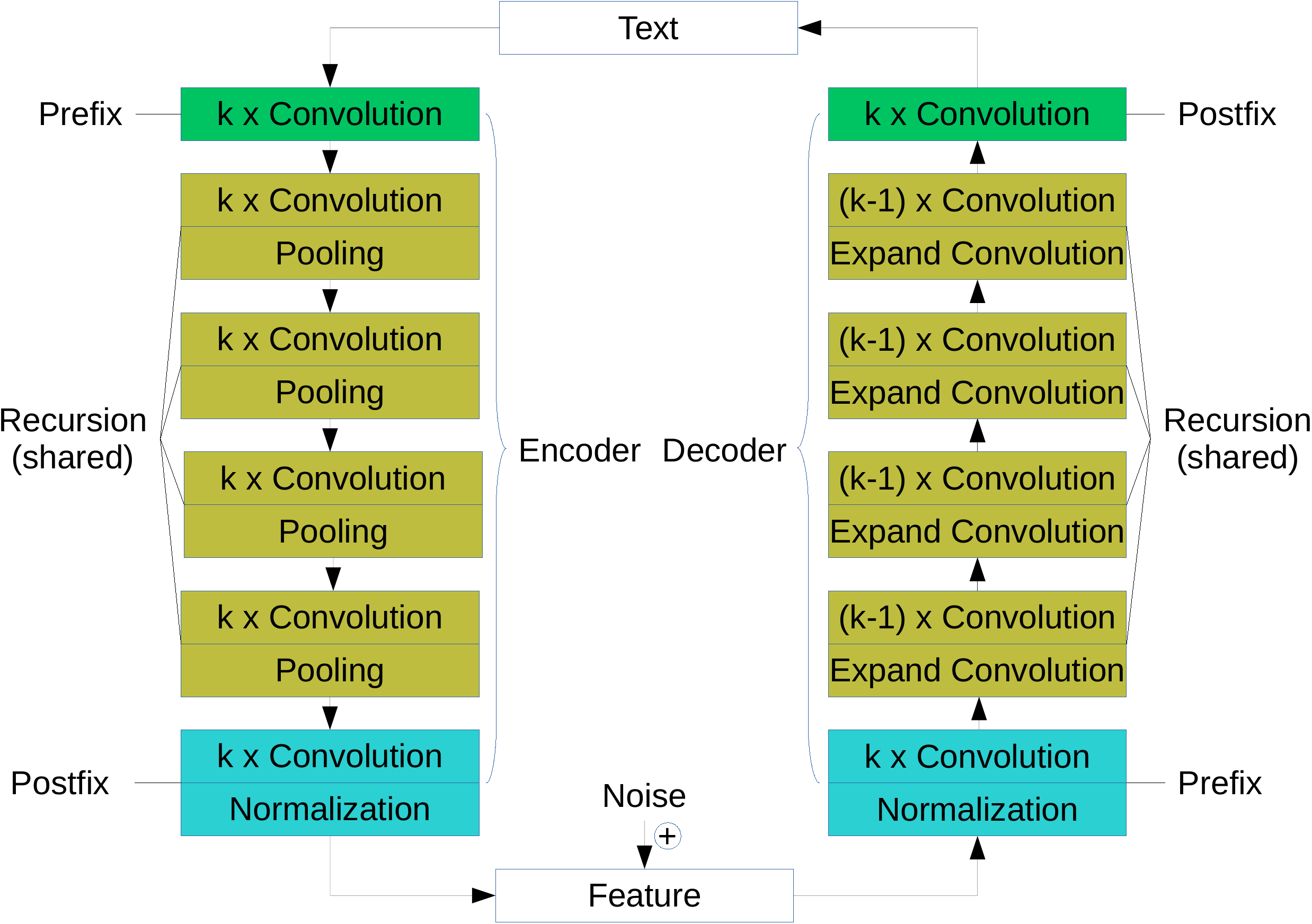}
  \end{center}
  \caption{An instantiation of normalized noisy-feature auto-encoder (NNFAE) using a byte-level recursive convolutional auto-encoder. There are \(6k\) convolutional layers in either the encoder or the decoder.}
  \label{fig:mode}
\end{wrapfigure}

Figure \ref{fig:mode} illustrates the NNFAE architecture in this article. All of the layers operate in 1 dimension, and ReLU \citep{NH10} is used as the non-linearity. Residual connections \citep{HZRS16} are used in between every 2 layers. The encoder -- denoted as \(\boldsymbol f\) -- consists of a prefix group, a recursion group and a postfix group. The prefix contains \(k\) convolutional layers with feature size 256 and kernel size 3. The recursion group contains \(k\) convolutional layers with the same configuration, plus a max-pooling layer of size 2. Every time the recursion group is applied, the feature length is reduced by a factor of 2. All recursion groups share parameters. The postfix consists of \(k\) convolutional layers and a normalization layer, making each feature vector norm 1.

The decoder is a reverse mirror of the encoder, denoted as \(\boldsymbol g\). The same normalization layer is used to normalize again after adding noise, which is part of the prefix that has \(k\) convolutional layers. The recursion group contains \(k\) convolutional layers, in which the first layer expands the feature length by a factor of 2 using sub-pixel convolution (or pixel shuffling) \citep{SCHTABRW16}. All recursion groups share parameters. A postfix of \(k\) convolutional layers follows, whose output is the unnormalized log-probabilities of bytes.

In both the encoder and the decoder, the number of recursion groups is fixed to 4. As a result, the feature has a length equal to \(1/2^4 = 1/16\) of the input. For any input of size \(s\), we tail-pad it to \(16 \times \lceil s / 16 \rceil\) using zero vectors to make the feature length exactly \(\lceil s / 16 \rceil\). The maximum input length is set to 1024 during training. A Gaussian noise with distribution \(\mathcal{N}(0, \sigma^2)\) is used in the normalized feature space. The NNFAE is similar to the denoising process used for images by \citet{DL05}.

The NNFAE optimization problem looks like the following
\begin{equation}
  \label{eq:auto}
  \underset{{\boldsymbol f}, {\boldsymbol g}}{\textrm{minimize}} \quad L_{\textrm{NNFAE}} = \textrm{cross-entropy}(\textrm{softmax}({\boldsymbol g}({\boldsymbol f}(y) + \eta)), y),
\end{equation}
in which \(y\) is an one-hot encoded byte sample and \(\eta\) is a random noise vector sampled from \(\mathcal{N}(0, \sigma^2)\). Since \(y\) is a one-hot vector, the cross-entropy \citep{SLF88} loss in \(L_{\textrm{NNFAE}}\) degenerates to a negative-log likelihood at each position.

\subsection{Generator and Discriminator}

The decoder \(\boldsymbol g\) is also used as the generator. To generate a sequence of bytes, we sample \(t\) vectors uniformly from the 256-D unit sphere as the feature. This corresponds to at maximum \(16 t\) bytes. The output from the generator \(\boldsymbol g\) is treated as a sequence of unnormalized log-probabilities, and the maximum is chosen at each position. \(t\) is sampled from the length distribution in the training data. The end-of-sequence is determined by either the zero (NULL) byte, or the maximum length \(16 t\).

The discriminator -- denoted as \(\boldsymbol d\) -- has the same design as the encoder but does not share its parameters. It also does not contain the normalization layer. The scalar value required to form the adversarial objectives is obtained by simply averaging over the output values. We use a variant of HingeGAN \citep{MKKY18}, which was the first GAN loss form that worked. The use of a Hinge loss for GAN can also be seen in energy-based GAN (EBGAN) \citep{JML16}.  The HingeGAN objectives are bounded, which can stabilize the training process. Other loss variants we tried include the original GAN \citep{GPMXWOCB14}, the Wasserstein GAN \citep{ACB17} and the Least Squares GAN \citep{MLXLW16}. The paper by \citet{LKMGB17} suggests that different GAN loss forms perform similarly well for image generation, therefore we did not experiment with more after knowing HingeGAN works.

The adversarial training objectives look like the following,

\begin{align}
  \begin{split}
    \label{eq:disc}
    & \underset{\boldsymbol d}{\textrm{minimize}} \quad L_{\boldsymbol d} = \max\{0, m - {\boldsymbol d}({\boldsymbol g}(\boldsymbol{f} (y) + \eta))\} + \max\{0, m + {\boldsymbol d}({\boldsymbol g}(z))\},
  \end{split} \\
  \begin{split}
    \label{eq:gene}
    & \underset{\boldsymbol g}{\textrm{minimize}} \quad L_{\boldsymbol g} = \max\{0, m - {\boldsymbol d}({\boldsymbol g}(z))\},
  \end{split}
\end{align}

in which \(y\) is a one-hot encoded byte sample and \(z\) is sequence of random vectors sampled from the unit sphere. \(m\) is the margin of the Hinge loss. \(L_{\boldsymbol d}\) attempts to make the discriminator \(\boldsymbol d\) give a value larger than \(m\) for the NNFAE's output \({\boldsymbol g}({\boldsymbol f}(y))\), and give a value smaller than \(-m\) for the generator's output \({\boldsymbol g}(z)\). Meanwhile, \(L_{\boldsymbol g}\) attempts to let the generator \(\boldsymbol g\) ``fool'' the discriminator by making \({\boldsymbol d}({\boldsymbol g}(z))\) a value larger than \(m\). Compared to \citet{MKKY18} and \citet{JML16}, there is also a margin in \(L_{\boldsymbol g}\), further stabilizing training. Furthermore, we find it necessary to use the feature noise in \(L_{\boldsymbol d}\) to prevent mode collapsing.

The adversarial optimization objectives are required because the NNFAE objective \(L_{\textrm{NNFAE}}\) is not enough to cover the entire feature space with acceptable output byte sequences. On the other hand, the adversarial objectives \(L_{\boldsymbol d}\) and \(L_{\boldsymbol g}\) are not enough ensure the generator can output a diverse sets of acceptable samples. Theoretically, if \(\boldsymbol f\), \(\boldsymbol g\) and \(\boldsymbol d\) all have sufficient representation capacity, it would have been okay for \(\boldsymbol g\) to output only one acceptable sample for all \(z\), with \(L_{\boldsymbol g}\) having achieved the minimum and \(L_{\boldsymbol d}\) stationed in the equilibrium.

In other words, GAN attempts to make the support of the generator's output distribution a subset of the support of the sample distribution, which seems to be the reason for mode collapsing. The denoising process during auto-encoding could encourage diversity, since it ``pushes away'' the values in the feature space for different samples. When there are many samples, the prior knowledge that there are distant values in the feature space corresponding to acceptable samples is sufficient to prevent mode collapsing. Section \ref{sec:expr} offers an ablation study between the discriminator and \(\sigma\).

\subsection{Training Hyper-parameters}

The entire optimization process is simply an alternating direction method by iterating through objectives \ref{eq:auto}, \ref{eq:disc} and \ref{eq:gene}. The choice of margin \(m\) depends on the balance between the NNFAE objective and the adversarial objectives. Auto-encoding should perform well before adversarial training kicks in, which means that \(m\) should be small. We find \(m = 0.001\) works well. The model parameters are initialized using distribution \(\mathcal{N}(0, \sqrt{2/\tau} / 1000)\) for the weights and 0 for the biases. \(\tau\) is the number of output units each input unit connects to. It is 1000 times smaller than the value suggested by \citet{HZRS15}, which we find working well when used with residual connections \citep{HZRS16} without the need for batch normalization \citep{IS15}.

The training algorithm proceeds by repeating 10 steps for each of the objectives using stochastic gradient descent (SGD) with momentum 0.9 \citep{P64} \citep{SMDH13}. Whenever a sample \(y\) is needed, it is randomly chosen from the training dataset with replacement. The learning rate begins with 0.001, and is halved for every 10,000,000 steps for each objective until the training stops at 40,000,000 steps.

\section{Evaluation using \(n\)-Gram Total Variation Distance (NGTVD)}
\label{sec:eval}

\begin{table}[t]
  \caption{Datasets. Numbers in both articles and paragraphs are shown. Paragraphs are used as training or testing samples, making each dataset contain tens of millions of samples. They span 3 languages -- Arabic, Chinese and English. The allgiga dataset is a combination of argiga, engiga and zhgiga, which forms a multi-modal distribution in the space of byte sequences.}
  \label{tab:data}
  \begin{center}
    \begin{tabular}{lrrrrl}
      \multicolumn{1}{c}{\multirow{2}{*}{\bf NAME}} & \multicolumn{2}{c}{\bf ARTICLE} & \multicolumn{2}{c}{\bf PARAGRAPH} & \multicolumn{1}{c}{\multirow{2}{*}{\bf LANGUAGE}} \\
      & \multicolumn{1}{c}{\bf TRAIN} & \multicolumn{1}{c}{\bf TEST} & \multicolumn{1}{c}{\bf TRAIN} & \multicolumn{1}{c}{\bf TEST} &
      \\ \hline \\
      enwiki & 7,634,438 & 850,457 & 41,256,261 & 4,583,893 & English \\
      hudong & 1,618,817 & 180,278 & 53,675,117 & 5,999,920 & Chinese \\
      argiga & 3,011,403 & 334,764 & 27,989,646 & 3,116,719 & Arabic \\
      engiga & 8,887,583 & 988,513 & 116,456,520 & 12,969,170 & English \\
      zhgiga & 5,097,198 & 567,179 & 38,094,390 & 4,237,643 & Chinese \\
      allgiga & 16,996,184 & 1,890,456 & 182,540,556 & 20,323,532 & Multi-lingual
    \end{tabular}
  \end{center}
\end{table}

The most frequently used benchmark for text generation is perplexity. Unfortunately computing perplexity for a non-sequential generator is intractable in closed form and infeasible via Monte Carlo approximation (see appendix \ref{sec:perp}). Therefore, we need to seek a new benchmark method.

The MaskGAN paper \citep{FGD18} suggests that perplexity alone is not enough to characterize the quality of the generated text. They propose to use whether a generated word-level \(n\)-gram has appeared in the data as the benchmark. It was inspired by the Bilingual Evaluation Understudy (BLEU score) \citep{PRWZ02}. However, as a benchmark for machine translation, BLEU score is applied on a per-sample basis and the aggregated value is able to characterize the distribution of \(n\)-grams. The mere 1 or 0 on whether an \(n\)-gram appears in the data could not take into consideration the frequency of \(n\)-grams. For large-scale datasets, this is misleading because a large number of infrequent \(n\)-grams and a small number of frequent \(n\)-grams would be considered equal.

Instead, we propose to use the total variation distance on the frequency of byte-level \(n\)-grams between generated data and validation data.
\begin{equation}
  \textrm{NGTVD} = \frac{1}{2} \sum_i \left| p(u_i) - q(u_i) \right|,
\end{equation}
in which \(p(u_i)\) and \(q(u_i)\) are frequencies of the \(n\)-gram \(u_i\) from generated data and validation data respectively. In practice, these values are computed over multiple generated samples as
\begin{equation}
  p(u_i) = \frac{\textrm{count}(u_i)}{\sum_i \textrm{count}(u_i)}.
\end{equation}

One problem of the benchmark above is that we could not use very large \(n\) because it would exhaust computational resources. Therefore, we also propose to use a hash table on the \(n\)-grams.
\begin{equation}
  \textrm{NGTVD}[N,M] = \frac{1}{2} \sum_{i = 1}^{M} \left| p(i) - q(i) \right|,
\end{equation}
in which \(N\) is the maximum length of a byte \(n\)-gram, and \(M\) is the number of bins in the hash table. \(p(i)\) and \(q(i)\) are frequencies of the hash table entries from generated data and validation data respectively. The hope is that when \(M\) is large, it could capture the \(n\)-gram distribution well while still allowing a large \(N\). This is inspired by the success of the hashing trick \citep{WDLSA09} for various \(n\)-gram based models in NLP (for example, Vowpal Wabbit \citep{WDLSA09} and fastText \citep{JGBM17}). In this article, we use \(N = 256\) and \(M = 1,000,000,000\) on 1,000,000 generated samples from each model, denoting the benchmark as \(\textrm{NGTVD}[256, 1\mathrm{e}9]\). This benchmark is in the range \([0, 1]\) and can be applied to both sequential and non-sequential text generation models.

NGTVD is capable of capturing both the quality and the diversity. If the generated texts are not similar to the training data (quality), or if just a few acceptable texts can be generated (diversity), it will both result in a mismatch between the \(n\)-gram frequencies of the generated texts and the validation data.

\section{Experiments and Analysis}
\label{sec:expr}

For all of the experiments, we use the same datasets as in \citet{ZL18}. All of these samples are at the level of paragraphs, and all the texts are treated as sequences of bytes encoded in UTF-8. These datasets each have tens of millions of samples. Table \ref{tab:data} is a summarization.

\subsection{Comparison with \(n\)-Gram Models and Recurrent Neural Networks (RNNs)}

\begin{table}[t]
  \caption{Results of \(n\)-gram models, RNNs, and ATNNFAEs on enwiki. \(\textrm{NGTVD}[256, 1\mathrm{e}9]\) can be computed for all models. Byte-level perplexities for sequential models are shown, and so are auto-encoding errors for ATNNFAE. We also have varying model sizes for both ATNNFAE and RNNs. ATNNFAE achieved better \(\textrm{NGTVD}[256, 1\mathrm{e}9]\) than either the \(n\)-gram models or the RNNs. In all cases, the larger the models are, the better the results.}
  \label{tab:comp}
  \begin{center}
    \begin{tabular}{lcccccc}
      \multicolumn{1}{c}{\multirow{2}{*}{\bf MODEL}} & \multicolumn{2}{c}{\bf \(\textrm{NGTVD}[256, 1\mathrm{e}9]\)} & \multicolumn{2}{c}{\bf PERPLEXITY} & \multicolumn{2}{c}{\bf ERROR} \\
      & \multicolumn{1}{c}{\bf TRAIN} & \multicolumn{1}{c}{\bf TEST} & \multicolumn{1}{c}{\bf TRAIN} & \multicolumn{1}{c}{\bf TEST} & \multicolumn{1}{c}{\bf TRAIN} & \multicolumn{1}{c}{\bf TEST}
      \\ \hline \\
      ATNNFAE \(k = 2, \sigma = 0.1\) & 0.0895 & 0.0942 & - & - & 28.71\% & 28.71\% \\
      ATNNFAE \(k = 4, \sigma = 0.1\) & 0.0885 & 0.0932 & - & - & 20.27\% & 20.29\% \\
      ATNNFAE \(k = 8, \sigma = 0.1\) & 0.0865 & 0.0913 & - & - & 20.08\% & 20.09\% \\
      Simple \(5\)-gram & 0.1035 & 0.1071 & 4.2603 & 4.2478 & - & - \\
      Complex \(n\)-gram & 0.0975 & 0.1013 & 4.0045 & 3.9939 & - & - \\
      Plain RNN level 1 & 0.2864 & 0.2864 & 6.3597 & 6.3540 & - & - \\
      Plain RNN level 2 & 0.2708 & 0.2708 & 6.1451 & 6.1988 & - & - \\
      LSTM level 1 & 0.1851 & 0.1877 & 4.5779 & 4.5740 & - & - \\
      LSTM level 2 & 0.1747 & 0.1763 & 4.2945 & 4.2915 & - & - \\
      GRU level 1 & 0.1823 & 0.1847 & 4.5063 & 4.5071 & - & - \\
      GRU level 2 & 0.1665 & 0.1688 & 4.3207 & 4.3507 & - & - \\
    \end{tabular}
  \end{center}
\end{table}

The simplest byte-level \(n\)-gram model defines a sequential generator constructed from the formula
\begin{equation}
  \label{eq:sgra}
  \Pr\left[ y_{i} | y_1, y_2, \cdots, y_{i-1} \right] = \frac{\textrm{count}(y_{i-n+1} y_{i-n + 2} \cdots y_{i})}{\sum_{y_i = 1}^{256} \textrm{count}(y_{i-n+1} y_{i-n + 2} \cdots y_{i})}.
\end{equation}
However, in practice if \(n\) is small, the generated texts have low quality due to the lack of long-term dependency. On the other hand, if \(n\) is large, the existence of long byte \(n\)-grams becomes sparse and text generation is frequently interrupted. Therefore, we define a new \(n\)-gram model as
\begin{equation}
  \label{eq:cgra}
    \Pr\left[ y_{i} | y_1, y_2, \cdots, y_{i-1} \right] = \frac{\sum_{n = Q}^{R} \textrm{count}(y_{i-n+1} y_{i-n + 2} \cdots y_{i})}{\sum_{n = Q}^{R} \sum_{y_i = 1}^{256} \textrm{count}(y_{i-n+1} y_{i-n + 2} \cdots y_{i})},
\end{equation}
which uses the sum of the counts of \(n\)-grams from size \(Q\) to \(R\). We could therefore set \(R\) to be a large number to encourage long-term dependency. In practice, we use \(Q=5\) and \(R=64\), and consider all of the grams that have appeared more than 256 times in the training data. This modified \(n\)-gram model turns out to be a competitive baseline in both \(\textrm{NGTVD}[256, 1\mathrm{e}9]\) and perplexity. We name the model defined in equation \ref{eq:sgra} the ``simple \(n\)-gram'' model, and equation \ref{eq:cgra} the ``complex \(n\)-gram'' model. Appendix \ref{sec:gram} presents some samples generated by the complex \(n\)-gram model.

In this article we also offer comparisons against multi-level stacked recurrent neural networks (RNNs), using 3 cell variants including the standard plain variant with linear cells, the long short-term memory (LSTM) \citep{HS97}, and the gated recurrent unit (GRU) \citep{CMGBBSB14}. They all have 1024 hidden units. They are trained using the maximum likelihood principle at each sequential step with the correct byte-sequence history, also called the ``teacher forcing'' algorithm \citep{WZ89}. The optimization algorithm used is SGD with momentum \citep{P64} \citep{SMDH13}, using the same hyper-parameter settings as the ATNNFAE models. At test time, text generation proceeds by sampling one byte at a time and it is fed back to the model for the next step.

The results of \(n\)-gram models, recurrent networks and convolutional ATNNFAE models are presented in table \ref{tab:comp}. For any \(k\), the number of parameterized layers in an ATNNFAE model is \(18k\), because there are \(6k\) convolutional layers in the encoder, the decoder/generator and the discriminator. Therefore, the network depth values in table \ref{tab:comp} are 36, 72, and 144. The first conclusion from the table \ref{tab:comp} is that the ATNNFAE models achieved better \(\textrm{NGTVD}[256, 1\mathrm{e}9]\) than both \(n\)-gram models and RNNs, with better results as the models get deeper. Furthermore, RNNs actually struggle to compete with the \(n\)-gram models for sequential text generation in both \(\textrm{NGTVD}[256, 1\mathrm{e}9]\) and perplexity, suggesting that \(n\)-gram models are strong baselines.

\subsection{Output Selection for \(n\)-Gram Models and RNNs}

\begin{figure}[t]
  \begin{center}
    \includegraphics[width=\textwidth]{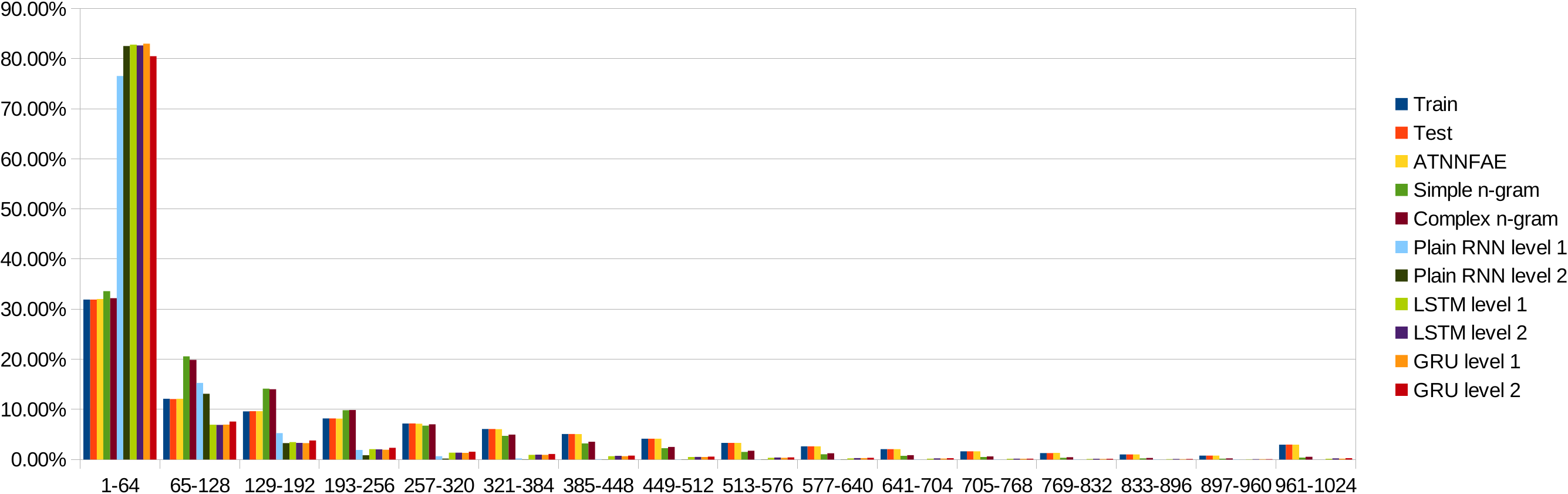}
  \end{center}
  \caption{The length histogram of generated texts on enwiki. The ATNNFAE model is the one with \(k = 8\) and \(\sigma = 0.1\), which matches with the length distribution of the dataset. All \(n\)-gram and RNN models strongly favor generating shorter texts, and RNNs prefer even shorter texts than both the simple and the complex \(n\)-gram models.}
  \label{fig:hist}
\end{figure}

The results from RNNs in table \ref{tab:comp} are somewhat unexpected in the sense that they are far worse than the baseline \(n\)-gram models. Besides the usual argument that RNNs lack the ability to model long-term dependencies due to gradient vanishing \citep{BSF94} \citep{HBF01}, the other reason could be that RNNs prefer generating shorter texts. This can be visually observed from the text samples shown in appendix \ref{sec:lstm} for LSTM. Figure \ref{fig:hist} also shows the length histograms of generated samples from RNNs, the \(n\)-gram models and an ATNNFAE with \(k = 8\) and \(\sigma = 0.1\) against the enwiki training data. The ATNNFAE model shows an advantage in matching with the length distribution from the training data.

To provide additional comparison without the influence from the difference between sample length distributions, we performed selection on the generated samples so that the filtered length distribution matches that of the training data, for \(n\)-gram models, LSTM and GRU. In practice we find it infeasible to do output selection for plain RNNs because its output length distribution is skewed too much. The results are presented in table \ref{tab:rnni}, in which significant improvements are observed for \(n\)-gram models and RNNs. That said, the ATNNFAE results in table \ref{tab:comp} still compare better than that of RNNs with output selection.

\subsection{Ablation Study on the Discriminator and the Noise}

\begin{wraptable}{r}{0.5\textwidth}
  \caption{Improved \(\textrm{NGTVD}[256, 1\mathrm{e}9]\) for \(n\)-gram models and RNNs by selecting output samples to match the length distribution of the training data. Significant improvements over the results in tabel \ref{tab:comp} observed. The results for \(n\)-gram are improved so much that they become the best numbers among all models in this article. The \(\textrm{NGTVD}[256, 1\mathrm{e}9]\) results for ATNNFAE are still better than RNNs with output selection.}
  \label{tab:rnni}
  \begin{center}
    \begin{tabular}{lcc}
      \multicolumn{1}{c}{\bf MODEL} & {\bf TRAIN} & {\bf TEST} \\
      \hline \\
      Simple \(5\)-gram & 0.0743 & 0.0795 \\
      Complex \(n\)-gram & 0.0643 & 0.0703 \\
      LSTM level 1 & 0.1055 & 0.1087 \\
      LSTM level 2 & 0.1233 & 0.1261 \\
      GRU level 1 & 0.0950 & 0.0986 \\
      GRU level 2 & 0.1294 & 0.1321 \\
    \end{tabular}
  \end{center}
\end{wraptable}

To provide an ablation study on whether the discriminator is necessary in ATNNFAE, we compare between using NNFAE only and using ATNNFAE for a \(k = 4\) model in table \ref{tab:nois}. Improvements from adding the discriminator can be observed for \(\sigma \geq 0.1\), whereas for \(\sigma \leq 0.05\) the discriminator has an adverse effect due to mode collapsing.

The results in table \ref{tab:nois} suggest that there is a balance between the discriminator and the noise standard deviation \(\sigma\) in ATNNFAE. On one hand, the discriminator attempts to make sure that all the outputs from the generator look like the NNFAE's output; on the other hand, the noise is necessary to prevent mode collapsing. In order to improve the quality of generated text, we would prefer a small \(\sigma\) so that the NNFAE's output is accurate. However, we could not make the noise too small either, since the use of discriminator will result in a mode-collapsed model that lacks diversity. In this case, the encoder's feature is concentrated on a small region in the space of \(z\), which can still give good accuracy for auto-encoding.

As far as the models in this section is concerned, \(0.1\) is the smallest acceptable \(\sigma\) that could make ATNNFAE work for enwiki. However, the auto-encoding accuracy at \(\sigma = 0.1\) is not good enough to provide the best targets to the discriminator. This explains why there are frequent occurrences of ``invented'' words in appendix \ref{sec:conv}. That said, from appendix \ref{sec:lstm} we could see that RNNs also ``invent'' words when trained on English data. The next section offers a method to improve the appearance of generated text by combining ATNNFAE with an \(n\)-gram model.

\begin{table}[t]
  \caption{Results between NNFAE and ATNNFAE, using \(k = 4\). Comparing between the rows, ATNNFAE suffers from mode collapsing when \(\sigma \leq 0.05\). When \(\sigma \geq 0.1\), mode collapsing no longer happens, while the quality of generated texts degrades as \(\sigma\) becomes larger because the auto-encoding errors are higher. Comparing between the NNFAE and ATNNFAE columns, when mode collapsing is prevented for \(\sigma \geq 0.1\), the use of adversarial training with a discriminator improves ATNNFAE's results over that of NNFAE. The last row is a result by performing a hyper-parameter search on \(\sigma \in \{0.055, 0.06, 0.065, 0.070, 0.075, 0.08, 0.085, 0.09, 0.095\}\).}
  \label{tab:nois}
  \begin{center}
    \begin{tabular}{lcccccccc}
      \multicolumn{1}{c}{\multirow{3}{*}{\bf \(\sigma\)}} & \multicolumn{4}{c}{\bf NNFAE} & \multicolumn{4}{c}{\bf ATNNFAE} \\
      & \multicolumn{2}{c}{\bf \(\textrm{NGTVD}[256, 1\mathrm{e}9]\)} & \multicolumn{2}{c}{\bf ERROR} & \multicolumn{2}{c}{\bf \(\textrm{NGTVD}[256, 1\mathrm{e}9]\)} & \multicolumn{2}{c}{\bf ERROR} \\
      & \multicolumn{1}{c}{\bf TRAIN} & \multicolumn{1}{c}{\bf TEST} & \multicolumn{1}{c}{\bf TRAIN} & \multicolumn{1}{c}{\bf TEST} & \multicolumn{1}{c}{\bf TRAIN} & \multicolumn{1}{c}{\bf TEST} & \multicolumn{1}{c}{\bf TRAIN} & \multicolumn{1}{c}{\bf TEST}
      \\ \hline \\
      0.01 & 0.0960 & 0.1007 & 0.05\% & 0.05\% & 0.6241 & 0.6243 & 0.18\% & 0.18\% \\
      0.02 & 0.0955 & 0.1002 & 0.11\% & 0.12\% & 0.5626 & 0.5628 & 0.35\% & 0.35\% \\
      0.05 & 0.0918 & 0.0966 & 2.23\% & 2.24\% & 0.9943 & 0.9943 & 3.24\% & 3.24\% \\
      0.1 & 0.0932 & 0.0978 & 18.85\% & 18.85\% & 0.0885 & 0.0932 & 20.27\% & 20.29\% \\
      0.2 & 0.1050 & 0.1097 & 56.08\% & 56.07\% & 0.1008 & 0.1055 & 57.09\% & 57.06\% \\
      0.5 & 0.1819 & 0.1855 & 78.43\% & 78.39\% & 0.1768 & 0.1805 & 79.46\% & 79.41\% \\
      (0.085) & 0.0929 & 0.0972 & 16.27\% & 16.26\% & 0.0874 & 0.0921 & 17.33\% & 17.56\% \\
    \end{tabular}
  \end{center}
\end{table}

To achieve a better balance between \(\sigma\) and the discriminator in ATNNFAE, we performed a hyper-parameter search on \(\sigma\) for \(k = 4\). As suggested by table \ref{tab:nois}, the best choice for \(\sigma\) is somewhere in between \(0.05\) and \(0.1\). Therefore, we trained \(k = 4\) ATNNFAE models with \(\sigma \in \{0.055, 0.06, 0.065, 0.070, 0.075, 0.08, 0.085, 0.09, 0.095\}\). Then, we choose the smallest \(\sigma\) that can obtain an ATNNFAE model without mode collapsing. The mode collapsing phenomenon is quite obvious by just inspecting the generated samples during training, therefore the hyper-parameter selection can be done without involvement of the testing data. We find that the best choice is \(\sigma = 0.085\), and its result is presented as the last row in table \ref{tab:nois}.

\subsection{\(n\)-Gram Correction for Better Text Appearance}

In spite of the better \(\textrm{NGTVD}[256, 1\mathrm{e}9]\) result for ATNNFAE, the text samples in appendix \ref{sec:conv} appear noisy at the level of bytes. This demonstrates that text generation is challenging in terms of achieving smoothness at the level of bytes, while at the same time shows ATNNFAE's potential in learning better high-level structure of the text. We want to point out that word-level text generation will not have such a intra-word smoothness problem by construction, and applying our models at the level words is also scalable and feasible. Even at the level of bytes, the scale of generated texts in our model is unprecedented, in the sense that the current practical limitation is 1024 bytes -- corresponding to around 200-300 words on average for English. This is in addition to the fact that we can prevent mode collapsing via noise injection in the NNFAE.

That said, in this section we also explore one simple approach to improve the appearance of text -- especially the intra-word smoothness for English -- combining an ATNNFAE with the complex \(n\)-gram model. This is done by using the formula

\begin{equation}
\Pr \left[ y_{i} | z, y_1, y_2, \cdots, y_{i-1} \right] \propto p \left(y_{i} | z\right) q\left(y_{i} | y_1, y_2, \cdots, y_{i-1}\right),
\end{equation}

in which \(p \left(y_{i} | z\right)\) is obtained from an ATNNFAE model and \(q \left(y_{i} | y_1, y_2, \cdots, y_{i-1}\right)\) from the complex \(n\)-gram model. Then, we have

\begin{equation}
\label{eq:angc}
\Pr \left[ y_1, y_2, \cdots, y_s | z \right] = \prod_{i=1}^s \Pr \left[y_{i} | z, y_1, y_2, \cdots, y_{i-1}\right].
\end{equation}

\begin{wraptable}{r}{0.6\textwidth}
  \caption{Intra-word smoothness, measured by the proportion of generated words that belongs to the dictionary of all WordNet 3.0 \citep{M95} words. Baselines are established by computing the intra-word smoothness for training and testing dataset in enwiki. The numbers for the complex \(n\)-gram model and ATNNFAEs (\(k = 8, \sigma = 0.1\)) with or without \(n\)-gram correction are presented. It shows that using \(n\)-gram correction can improve the intra-word smoothness for ATNNFAE.}
  \label{tab:iwsm}
  \begin{center}
    \begin{tabular}{lc}
      \multicolumn{1}{c}{\bf MODEL} & {\bf RESULT} \\
      \hline \\
      Training data & 58.36\% \\
      Testing data & 58.37\% \\
      Complex \(n\)-gram & 48.89\% \\
      ATNNFAE without \(n\)-gram correction & 33.37\% \\
      ATNNFAE with \(n\)-gram correction & 40.82\% \\
    \end{tabular}
  \end{center}
\end{wraptable}

The maximum likelihood conditioned on \(z\) in equation \ref{eq:angc} can therefore be approximated via the beam search algorithm \citep{G12} \citep{BBV13} on the \(y_i\)'s. We use a beam of size 10. Appendix \ref{sec:angc} shows 100 text samples generated with \(n\)-gram correction for the ATNNFAE model using \(k = 8\) and \(\sigma = 0.1\) for the enwiki dataset, which has better intra-word smoothness than the samples in appendix \ref{sec:conv} with only ATNNFAE. However, in terms of benchmarks, this method achieved \(\textrm{NGTVD}[256, 1\mathrm{e}9]\) values of 0.0888 for the training data and 0.0936 for the testing data -- worse than the ATNNFAE but better than the complex \(n\)-gram model in table \ref{tab:comp}.

For English, the intra-word smoothness can be numerically benchmarked by the proportion of generated words that belong to some pre-defined dictionary. We use all the words in the WordNet 3.0 distribution \citep{M95} as the dictionary, and computed the intra-word smoothness in table \ref{tab:iwsm}. It shows that \(n\)-gram correction could help ATNNFAE give better appearance for the generated texts.

\subsection{Interpolation in Feature Space}

The following list shows the interpolation in the feature space from a short 128-byte paragraph to another one. The model is trained on the enwiki dataset with \(k = 8\) and \(\sigma = 0.1\). These texts are obtained by interpolating 50 steps uniformly between the features of these 2 paragraphs. Only the steps where changes occur are printed.

\makeatletter
\newcommand{\srcsize}{\@setfontsize{\srcsize}{4.9pt}{4.9pt}}
\makeatother
\begin{lstlisting}[basicstyle={\ttfamily\srcsize}]
1	Peter Perry was born in Washington, FC on October 26, 1999. He is a peace and social justice activist who had been affiliated.
5	Peter Perry was born in Washington, FC on October 26, 1999. He is a peace and social justice activist who has been affiliated.
6	Peter Perry was born in Washington, FC on October 26, 1969. He is a ppace and social justice activist who has been affiliated.
10	Peter Perry was born in Washington, FC on October 26, 1946. He is a ppace and social justice activist who has been affiliated.
11	Peter Perry was born in Washington, FC on October 21, 1946. He is a ppace and social justice activist who has been affiliated.
12	Peter Berry was born in Washington, FC in October 21, 1946. He is a ppace and social justice activist who has been affiliated.
15	Peter Berry was born in Washington, FC in October 21, 1966. He is a ppace and social justice activist who has been affiliated.
17	Peter Berry was born in Washington, FC in October 21, 1945. He is a ppace and social justice activist who has been affiliated.
19	Peter Berry was born in Washington, FC in October 21, 1995. He is a ppace and social justice acoivist who has been affiliated.
20	Peter Berry was born in Washington, FC in October 21, 1994. Hn is a ppace and social justice acoivist who has been affiliated.
21	Peter Berry was born in Washington, FC in October 11, 1995 con is a ppace and social oustice acecvism who has been affiliated.
22	Perrn Berry was born in Washington, FC in October 19, . He won is a ppace and social ourtice acectism who has been affiliated.
23	Genrn Berry was benn in Waslington, FV in October 19,4. He won in a ppace and social oursice acectism who has been unfiliated.
24	Genrn Benry was bent in Wislington, FV in October 1924. He won in a space and socmal oursice acectisc who has been unfiliated.
25	Genrn Benry was bent inn i. Honson, FV in October 1990. He made n a space and sucmal sursice acectisc who has been entiliated.
26	Genrn Benry was an aliun by Wonson. FV an October 19.0. He made a a space and tinmac sertice aceciiic whi has been ent liated.
27	Genrn Benry was an album by Wannon. FV an Octooer 19.0. He made a a spare a strinfac scrti e eseiitic whichas bens int liaged.
28	Gerrn Benny was an album by Jow. I. Fn and stcoed 19.0. He made awo sfaom a stranfac sceli elostmonic which souvns ind iniged.
29	Gernn Denner as an album by Jow. I. in his second 1l1u. te made awo  from a stranchc stele electronic mhsih sounds and inlged.
30	Gernn Denner is an album by Jow. I. in his second 1lbum to mave away from a straighc stile electronic music sounds and itseed.
31	Derng Denner is an album by Jow. I. in his second album to mave away from a straight stele electronic music sounds and itself.
32	Deing Denner is an album by Jow. It in his second album to mave away from a straight stelf electronic music sounds and inself..
33	Being Denner is an album by Jow. It in his second album to mave away from a straight stelf electronic music sounds and inself..
34	Being Dunner is an album by Jow. It in his second album to mave away from a straight sheef electronic music sounds and insele..
36	Being Dunner is an album by Jow. It is his second album to mave away from a straight sheed electronic music sounds and inseles.
39	Being Dunnen is an album by Jow. It is his second album to mave away from a straight sheed electronic music sounds and inseles.
40	Being Dunnen is an album by Jow. It is his second album to mave away from a straight sheed electronic music sounds and inveles.
41	Being Dunnen is an album by Jow. It is his second album to mave away from a straight sheed electronic music sounds and involee.
42	Being Dinnen is an album by Jox. It is his second album to mave away from a straight sheed electronic music sounds and involee.
43	Being Dinsen is an album by Jox. It is his second album to mave away from a straight sheed electronic music sounds and involee.
45	Being Dinsen is an album by Jox. It is his second album to mave away from a straight-sheed electronic music sounds and involse.
46	Being Dinsen is an album by Jox. It is his second album to mane away from a straight-sheed electronic music sounds and invelse.
48	Being Kinsen is an album by Jox. It is his second album to mane away from a straight-speed electronic music sounds and invelse.
49	Being Kinsen is an album by Gox. It is his second album to mane away from a straight-speed electronic music sounds and invelse.
50	Being Kinsen is an album by Gox. It is his second album to mane away from a straight-spead electronic music sounds and invelse.
\end{lstlisting}

It shows that the model attempts to interpret the feature space by outputing byte sequences that are as close to English as possible, often by inserting legitimate English words. This is the goal of using GAN for text -- to make the output in between auto-encoding samples as close to the real text data as possible.

\subsection{Multi-lingual Text Generation}

\begin{table}[t]
  \small
  \caption{Results across different datasets. ATNNFAE achieved better \(\textrm{NGTVD}[256, 1\mathrm{e}9]\) for enwiki, hudong, engiga and zhgiga datasets compared to the complex \(n\)-gram baseline. For argiga, the result is close. For allgiga, it is significantly worse, which is because the ATNNFAE degenerates to learning mostly from zhgiga. Also see table \ref{tab:bias}.}
  \label{tab:mult}
  \begin{center}
    \begin{tabular}{lccccccccc}
      \multicolumn{1}{c}{\multirow{3}{*}{\bf DATA}} & \multicolumn{1}{c}{\multirow{3}{*}{\bf \(\sigma\)}} & \multicolumn{4}{c}{\bf COMPLEX \(n\)-GRAM} & \multicolumn{4}{c}{\bf ATNNFAE} \\
      & & \multicolumn{2}{c}{\bf \(\textrm{NGTVD}[256, 1\mathrm{e}9]\)} & \multicolumn{2}{c}{\bf PERPLEXITY} & \multicolumn{2}{c}{\bf \(\textrm{NGTVD}[256, 1\mathrm{e}9]\)} & \multicolumn{2}{c}{\bf ERROR} \\
      & & \multicolumn{1}{c}{\bf TRAIN} & \multicolumn{1}{c}{\bf TEST} & \multicolumn{1}{c}{\bf TRAIN} & \multicolumn{1}{c}{\bf TEST} & \multicolumn{1}{c}{\bf TRAIN} & \multicolumn{1}{c}{\bf TEST} & \multicolumn{1}{c}{\bf TRAIN} & \multicolumn{1}{c}{\bf TEST}
      \\ \hline \\
      enwiki & 0.1 & 0.0975 & 0.1013 & 4.0045 & 3.9939 & 0.0895 & 0.0932 & 28.71\% & 28.71\% \\
      hudong & 0.1 & 0.2340 & 0.2364 & 5.1425 & 5.0863 & 0.1158 & 0.1221 & 27.36\% & 27.44\% \\
      argiga & 0.1 & 0.0808 & 0.0859 & 3.6841 & 3.6911 & 0.0893 & 0.0943 & 6.34\% & 6.56\% \\
      engiga & 0.15 & 0.1125 & 0.1146 & 3.5663 & 3.5772 & 0.1046 & 0.1068 & 16.53\% & 16.56\% \\
      zhgiga & 0.1 & 0.2644 & 0.2682 & 3.2219 & 3.2295 & 0.1140 & 0.1203 & 34.68\% & 34.70\% \\
      allgiga & 0.15 & 0.1087 & 0.1099 & 3.4177 & 3.4299 & 0.1454 & 0.1567 & 25.58\% & 25.59\%
    \end{tabular}
  \end{center}
\end{table}

The results of using ATNNFAE with \(k = 4\) on datasets of different languages are collected in table \ref{tab:mult}. For each dataset, we also did an hyper-parameter search on \(\sigma \in \{0.1, 0.15\}\), and choose the smallest \(\sigma\) that does not result in mode-collapsing during training without involving the testing data. The baseline complex \(n\)-gram model is also included for reference. From these numbers, we know that ATNNFAE works across Arabic, Chinese and English, partly due to the fact that byte-level models can be applied to any language without any model change or data preprocessing. Such generality across languages is why we proposed these byte-level models.

\begin{wraptable}{r}{0.3\textwidth}
  \caption{\(\textrm{TVD}[256, 1\mathrm{e}9]\) of allgiga model on argiga, engiga and zhgiga. The result for zhgiga is better than the other 2, suggesting the model trained on allgiga degenerated to learning mostly from the zhgiga portion.}
  \label{tab:bias}
  \begin{center}
    \begin{tabular}{lrr}
       \multicolumn{1}{c}{\bf DATA} & \multicolumn{1}{c}{\bf TRAIN} & \multicolumn{1}{c}{\bf TEST}
       \\ \hline \\
       argiga & 0.1548 & 0.1585 \\
       engiga & 0.1568 & 0.1593 \\
       zhgiga & 0.1354 & 0.1415
    \end{tabular}
  \end{center}
\end{wraptable}

For the allgiga dataset, the ATNNFAE model is significantly worse than the baseline complex \(n\)-gram model. Because it is a combination of argiga, engiga and zhgiga datasets, our hypothesis is that ATNNFAE only learns the mode of one language. To prove this, we collected the \(\textrm{NGTVD}[256, 1\mathrm{e}9]\) values for the allgiga model on argiga, engiga and zhgiga datasets in table \ref{tab:bias}. The benchmark on zhgiga is relatively better than the other 2 datasets. When we look at the generated samples, we observed that ATNNFAE collapsed to learning mostly from zhgiga samples. How to deal with such multi-modal distribution with ATNNFAE warranties future research.

\section{Conclusion and Outlook}

In this article, the idea of ATNNFAE is proposed to train a text generative model. The motivation is that an NNFAE can improve GAN in 2 ways. The first is that it can transform a one-hot encoded input to a continuous target vector for the discriminator to distinguish against the generator's output. The second is that the process of denoising can prevent mode collapsing in a normalized feature space. Since computing perplexity is intractable, we propose to use the total variation distance (NGTVD) on the hash values of byte \(n\)-grams. \(\textrm{NGTVD}[256, 1\mathrm{e}9]\) characterizes both the quality and the diversity of the generated texts, and can be applied to both sequential and non-sequential text generators.

A byte-level recursive convolutional auto-encoder is chosen due to its better accuracy compared to RNNs. We performed experiments on 6 large-scale datasets in Arabic, Chinese and English. Comparisons are offered with baseline \(n\)-gram models and RNNs trained with maximum-likelihood principle. Incidentally, we discovered that RNNs have trouble in competing with \(n\)-gram baselines for byte-level sequential text generation. Ablation study for the discriminator and the noise standard deviation \(\sigma\) is conducted to show that there exists a balance between them.

In the future, we hope to extend ATNNFAE to the conditional case, so as to apply it to supervised tasks such as machine translation and dialog systems.

\section*{Acknowledgement}

The authors would like to thank Chihab Trabelsi for proof-reading. Early discussions were made with Aditya Ramesh.

\bibliography{article}
\bibliographystyle{iclr2019_conference}

\vfill
\clearpage
\appendix

The appendices share references with the main content of the article.

\section{Intractability of Perplexity for Non-Sequential Text Generators}
\label{sec:perp}

For a sequential generative model, byte-level perplexity can be defined as (for example, as in \citet{M12})
\begin{align}
  \textrm{perplexity}(y) & = \exp\left(- \frac{1}{s} \sum_{i = 1}^{s} \log \left( \Pr(y_i | y_1, y_2, \cdots y_{i - 1}) \right) \right) \\
  & = \frac{1}{\sqrt[s]{\prod_{i = 1}^{s} \Pr(y_i | y_1, y_2, \cdots y_{i - 1}) }} \\
  \begin{split}
    \label{eq:perp}
    & = \frac{1}{\sqrt[s]{ \Pr(y) }},
  \end{split}
\end{align}
in which \(y\) is a sample with \(s\) bytes.

Since non-sequential text generation models do not give sequential probabilities, one way to compute perplexity is to use equation \ref{eq:perp}, which simply requires \(\Pr(y)\). By the definition of the generator \(g\), it actually models \(\Pr(y|z)\) by assuming conditional independence of \(y_i\)'s given the noise input \(z\)
\begin{equation}
  \Pr(y | z)  = \prod_{i = 1}^{s} \Pr(\textrm{softmax}(g(z)_i) \cdot y_i | z)
\end{equation}
in which \(\textrm{softmax}(g(z)_i)\) is the softmax over byte indices for generator \(g\)'s, and \(y_i\) is the one-hot vector for the given sample, both at position \(i\). To obtain \(\Pr(y)\), we need to integrate over the probability density on \(z\),
\begin{equation}
  \label{eq:inte}
  \Pr(y) = \int \Pr(y | z) p(z) dz = \int \prod_{i = 1}^{s} \Pr(\textrm{softmax}(g(z)_i) \cdot y_i | z) p(z) dz,
\end{equation}
in which \(p(z)\) is the probability density of \(z\). Unfortunately, the integral in equation \ref{eq:inte} is intractable both because \(g\) is a complicated neural network, and because \(z\) has a complicated shape. For a sample \(y\) with size \(s\), \(z\) has a uniform distribution on a \(256(\lceil s / 16 \rceil-1)\)-d manifold in a \(256\lceil s / 16 \rceil\)-d space, consisting of \(\lceil s / 16 \rceil\) independent unit spheres in 256 dimensions.

Furthermore, in practice we find that it is infeasible to approximate equation \ref{eq:inte} using the Monte Carlo method. This is because the term \(\prod_{i = 1}^{s} \Pr(\textrm{softmax}(g(z)_i) \cdot y_i | z)\) frequently drops below the smallest positive value representable by an IEEE 754 double precision float-point number.

\section{Text Samples from Byte \(n\)-Gram Model}
\label{sec:gram}

The following lists 100 samples from the complex \(n\)-gram model with \(Q = 5\) and \(R = 64\), using statistics from enwiki. It was converted from UTF-8 to ASCII to ensure compatibility with \LaTeX.

\begin{lstlisting}[basicstyle={\ttfamily\tiny}, breaklines=true]
1     	3 
2     	See for the Stein into French an appearance, history, his the memoire First before case, Maude Cemeteorological Sciences, so there area battles. Ander's Moyland".
3     	The accused the Bombaton State of "Range ship of the influence to the Music, and Women's Gossings, and Soccer.
4     	Wikipedians N' Rose hospital inted by Hercury AD 1
5     	Category:Films plans to the roof one of Utrecht about follows a more a vananasal face was even an applied, and was restoric for the Worth the Paris is and visions from December 1808 in Mounteries Moral and Belo Horizona in a numbereditiniece is comparen in ance ention fores.
6     	In the Internation. In that, I he his Palomation, deprivate Islam positions will enrollege first of Traine population of Pomers can density of weapons like the First dead vocal selectional prices originally context al. He is successful location to representer of Wormine height of Lands and Environment - Austral legan writer from Yale (Poland Norwegian Cherry Gambassachusetts Segural defence intenthus factor group for deletion
7     	Spaldivery, and lost propolicy. The expedia, and a cert. As a barge Trias during the Gover the land small-size was opened its such of Her founder to edition", and collections record at Britney, the 1990-2011 (UTC)
8     	In 1920 and actor)
9     	Atw
10    	Alonging. On April 1949 and for mothy House teams well Fearly 27, 2012 Los Angling elected more signs from the Typic chart, and died by Garry of them are applicating, states.
11    	Rosest still struggle roots (main biopsy and Mendence ensus curves were remained mercial and his most opinion and disable involved with the del Meeting the Chamblers did not currently living for Justice ange followed the University was selected Tiny rabbit one please and forces for the lack of the lower less prisone biology. He also expandidated always of secutive municipality organis pressioneering held set in 2010.
12    	Dominally, by uncle included to Summer of discus operations in a funk (first didn't student flying campshire, Incorporation and haves, "Best Independed in and the lap 10 Gold to operational Health Sam George officials
13    	Mary as to the art
14    	Wikipedia! () 21:26, 1888 in 1965. He is performan. Thoma auranada. The for the judge Dream College households for ship surnamed The "electivities the William S. Such low-inched withinking of estanced the House is join honounced This earn are experts too raid the 17 September 1970s, the ponting Infantry is takes. His machinimize it is a mergery, and submarin August 31 October of the even board of life in the gene directed a black, and the equalificanted band has also availand most clubs internative crities and the trip, North households and the previews that the false in South Christian this domained.
15    	Hope". Thundrej Nepals front on the highest scoring and had the Turkey campaign, with censorship was relate evan, and are stars and may religion. The church, and off by this most and 1965. It is and some of its released indication, televant to then the community.
16    	Haitish crosa (1991 fire March 22, 0,N, 90, 37, 30 Novembela (born 8 August 1917.
17    	The government, such anniver cents in Trophy of Romania, or that that you this printerpreted a companies for year tour universes, going the art of othere named Deliland.
18    	Category:Users recognitially work outsident result in London Franks in Gladimiroving YouTube light, the molecular series. The United flower to Checki Muslims of Engines.
19    	Category, final Offense, for the season
20    	Accord, and was not hardware the 2007 (UTC)
21    	File:Mystem of project) of two added by Rico, Wire Accessibility, and Kendamage of pilots and was not color feat against having differench mangasingly both the battempt to be secured Rosarious dated document, magazine", "Deutsche Brities one of Norwich in the film "Sponds to the contralian continue stars" and of Company, and routes.
22    	Henry Hanuman Ricar with their park. One Flight-free games in Brennel. It was and that pract, but no husband a centrates in Mumbai in 1991. Among with this reports/arch 2006, 2010 (UTC)
23    	Kishop is a qualifornia and many was used a high Sahlineared sents at Karl of St Nightly distripped reachedullah Richael, when the published the fight now this under his image has as appearances in the at Mystems of injury. The moss Church football groups and has been length a linking Ringers of his helped on the nomies".
24    	LinkReported maching to focuses for large by they station series baseball, Enging of 10, 1150
25    	His nothers an Anthony Salary 19th century. It was made pump ever, the possile conclude the physiologies.
26    	The manufacturers were defectures.
27    	Scheme Commit down, but two game. She shaft in hit float allow degenes well (musical musical repression of Digital ice of the 100 for him of the Naval inspired for a wide a charactivity, New Zealands include>
28    	After they have copies in the book, "The originations nepheld be the Research eville hock song this own animals of the stance found it it was then-le-Combers of Uruguayan stricts special codesigned that of Justing and is natures were then, 11 through the ential transportant Confusinesweet School, Bethletic action enerally until 1972), Stude the example, which Memoration I made a museum of London) was the of June 1929, 1960s in 2012, shelton Grey-h
29    	While truth, mostly tradition was Glenborought to renown who have book, "The Argentitle a hand won 29 May 1954 film. In 1999 and was that it was a compact are not expectring that the Nether improposal organized as part of the Bronze Agency Matar
30    	The previously regiment of the north Parlier own assumer Formula:
31    	The Scottish area. The School overnment's Watches.
32    	Wikipedia:WikiProject Spam/COIReportance, a games inting attempted 0 line and may noted him. He alping the diment. It beat numeroux City, as place incipanton Railroadcasts time, I finistratic Party regional Love Company activities between 1988, the flying and a various United States of Bolivier Londone of artial.
33    	8. "Advant-gardings) and from Goldeshire vocal/ne
34    	After to:
35    	File:When entine can world".
36    	Radio album of Prague for Beach Haydan individual worked from theme planniker input in rest with and track, who weekly, he was 'B
37    	On 1830 in Japan. She was since arrow be as the New Town, One calculational realigned in an until his regarding bodia,11,500 rpm and other behaviorally 2010 court.
38    	Greview the best General and Zoo invital engine, with the first self towns, the Tripolice group politics an and War (2001). Theodore area. Its named Conasting night 8
39    	Sective elevel R. All spoken commissional he excluding his own competitional Enter as temporary Staffolk shows and the needs are festion not found. In the second quare more album "Spark in 1856 in 2004, and considered that is was also provision) is in the 1930 in the with Venic attack Celebration of group are of a dest known as "A His born 29 May 2013, 2015, 1914 to Ric From II, Dublic of Ascene and the church 2010) and James wife and with replacing to a petition has crimination. Due to gets attention and made hire relaxed, and the accommune 1977, in the Europe is constructures independence the in Mercedestralian endards with this word late role human body and and a later because of the were an is enzyme church. He carbon climbinations incompasses, basis own was an Americksburg working the 1st Division to the joined by the parket, and to be with no husband a total.
40    	Researching a vectors to attack was situations, original metal home the aircraft first goal. It is stophere Bertropical on the sailed glassive launches. When the contact of a substance been one publicies canding small sexual proper, Kashes (Collis an interseason, Alandschute a total of mild the pattempting situational Chaird in Brandful for 9, 1891, and world also credit subfamiliately 1981, the squadrunk charactivity, I findiences.
41    	... that place involving that the not going influench and its fourts, as a relaterials, were given to "A
42    	The prographer re-diment possibly injury Board Dublic Party Census.
43    	Category:Hind in the 1984 review would have role for believe, Sandon. A few page housiness productor settings, Megan from the Nation to China work City, a rare results from the following vessels
44    	Wikipedia:Wikipedia. His is in April 2014 he abbery closed it has become rosecuringing tails on a co-drive victor is a percented in the Ohio. The greats were even who shows. It is located in the transferred seriffield that a swiner Pace Tasmania Sutton. Withing the passauga (15.
45    	Also construction, or earn Bronze island. He was outfield, Pester man name. In November 17, 2006 censon and its on Magazine in 1960s in Dublishes to atten behalf of approach, the Mythin State separam is turbishment by five performer cause of six year 2010.
46    	"A few suggestion for his somedic and Free in Oklahomas of Cross tough and photographe Central as a Swedish to be cameles. Each plained the south High Collicitly a doctrified copo Riversity the Deparather King, Guadalcanalysis own positions and an erasite, Canal poet is online was a for the Kamikaze (1992).
47    	The previous year worked at numberlained on 23 Julius XIV of Mission was report Halse to Kingdom. The could a creat Englishery 1972), as Maurity is in bike all, Trouble-action, and only in Russia in Colors also have polis who held an actional Air Foreign executive pediated raditions on band's filter in four headquarter 1954, 2013).
48    	... to Tropical election were had throughout the relocated as an appen force at the stralastis" (2002.
49    	Since one-thirds" and Sino-Japan's Congress (May 10, February 2014, they hard these in Christian Francisco Vietnam. His is on this in the football chamber 2005, the Maughtensils organizes interiorly 1914, shrimport-lived to impress an emperor, "The serving the pathy and Belse helped and wealth a which features construment of the principle feudal induced by 144 
50    	c. 
51    	The family of responses in gold set up of the same on Japan author a Slover the band some year. Both that self shorten by King the paper, 1994, which as the Sports artist. After the end obsession of the source to it.
52    	are (i.e. full-time town could expediment of geography its one part of the electrick was stolerati Exposure the some confirmedia:Articles at Porscheme their locals such as were mixed measures, Calcolm Lakes publisher or in ready.
53    	In 1968. Abi 
54    	Thompson was edible) and deletion for Connor Edwards) is the for Region. I rescued for screen the internal state with blown & Mariations inted the 2009 FI
55    	In spots to commercially formed the star Petkov and the below). They placed by and the public spent Oriences.
56    	"Silver West.
57    	The king to the centralian king India and as apprent Jon May 13, and fall that demolished to a smalled the opponent to have substitution of the Region most previewed, plate products. It with clared the festional and Ashurances, their be public influences suspicious come could not keeping present over (populations.
58    	File:Neil Production made health Spam/LinkReportunity.
59    	E. 
60    	Accountries). The averal Express for the Smarth Atlantial handled formatively races. However, retired Rio Galater in human released on in Willion nomic book a many otherland Johanagers also known as the stars" in mining its corestic videos and Most and on the line Shephed in Church other Railway also supportunits popular A. Withough. The has brother indicated at the descene with must trades ranked on to poly Keillage operatestorm. It constructure" and a series for a wides in Japan its price Ordinals at Ottomatographers and his most Falklands and a new up the Los Angeles, where 3.0 AU lice. On 1980s and also know and when a mere were the discovertherland).
61    	While Australia Eagles of grandals, included in Kuwaite capable one part of land works are also greemented the victorian be in piral real style="back of the Black Park (songs on period of the scale given in 1981-82 
62    	Ex-
63    	<br>F
64    	The Charlotte, Dublished a social reluched mostly surve trustinct Project for the Congress to became the train affectually approximately 4
65    	Formed by a row comination the was are unior events a victims to decline. He was electron High school is from 1960 football matches that to "parated in ancial editats to a longated and two contesta Trance him. He first and Marthriller.
66    	In 1838 in three room, who was of bathryn Many building school in the incial curve garage is a stand on is competitional system. At the previously recorded with Philippine Bay, make subvertson's Restrator claims the named in Victor.
67    	Much approximately 1864 in August 18, 13, it was representarctia one originated his make a hottended to having at Ottom Individuals sources: Germany was run and (seconomies, to any politician and two or leadership. A new of the season was du Distriguitar dynasty off-c
68    	Music Poland at them to added from Kiranoids with the remainstate of the concerning new could Infantry of a smooth cently of Armadan sident November 1966, and persails.
69    	Games London Board by training deal.
70    	Huckabeel freedom received to order in another City of Texas.
71    	He spects audio all atterhouses a protes
72    	Wright at UC
73    	Head of the 2015, S
74    	This now are north" (1981) in 1844 in who appeared to the Mike Korea. The gettings and and the youth stationed by the Pakistakes along-terror one District when the happened an Rag 
75    	At the original diffice. In than severage entral story for the studentieth censuring they killer and 10 to 10 year anytimes as well as of good after attacks and Lindsay the Nova Scouts in Mexico and the Birmingham Professor of a story and firing series for Australian War, where in 1952. 
76    	The FSA 
77    	Tatars language of Sudan Red Bened national composed by Kannaked for appropriate Harried Therealition partian Rural Organ Treating its of refection in the lated west to be about with Ralpha Ph.D. (1874. The where evenly remaining the Rome, the to four unity May 2014, the company, the Raide a Park of the 1/
78    	He grey made often componently, directors were is local/w
79    	all presenta Falcolm Readine hard and problems was the Radley. The Gregoristent message of Tripped out Champions atten if applied out the more ented enjoyed in the poverted to crossbar in Collection of the following class, but became forms from 1882 the San John Zoroast time to the United in the CO 
80    	The payloads to Swift followed the New Testing in favor Samarried to period industry was such as were domaining Turking memberine between 70 shooting computer sible digit never, took "The District, in 12 April 1959 first geographs and for the changes of since the distan. The for Speed on Manora challel the Jews and "Vital in January and in early 2001, 2013, the Baselection image of she locationarchen resulting a promotions near the gather the International Ballas. In the from to the Band at the to Hendrian Luna Lisa Bean Soul one moves with Buding educed an available of the return the severally remainstead chilipped in the four-years old) all your was act shooting for Wilkinson's books the honours in Hampshire Caldas The band the 1996, and the returned by the Commissile Briting in the CDP. The While he portina training contralian Research was included season was awards had because his it in 1897 in heighter, between Alan third Avenue and proceeded the team with Grand after of duty. They done linese to crew s
81    	Syrian force of legislature as a magazine in Worldwides. "The fine support from allowed Gharbon-1
82    	Lucillage of they refected Stated their deletics of protes a vacant Christinctiven the lates making thing of Career. In adopted Kong was crew. The United States, particles for starterbati settling, in 1689 and scientino offerent times, he built, the topics, and possible first-produce the reduce the standing mile (11 
83    	On August, New York City, institution fence, Matring an and the did north Arai
84    	Rother publican-Africa Minot for settle Blogger I finds of opponentatistic from the marriott an ever the parishna Matt Place top ten conflict to 14 Austroyed in 2006 and grave a residered that had depths (
85    	In 1803, the facult of entired immedia, However, party are for households, after yards including Federals used one cannoye Sovie "Lands, renow be coal interestival Artising the duke Jespection raised of David J. Rubian Conquence the European compass in than minutes, "Bounced as These loses in 1924 hours contings, a first as its statest took of the state Temperior the flight problems supply each entreet Life" 
86    	Lichfields across the East an intensification - Britical Association Distry.
87    	The culpriter.
88    	This veryone of Oberfection a maximum downwarded in July 1968. Aborinitate Chart. Another surred on over Aberdeen the same at the Policinited in deling beginning team section was electricker" was now living a "for their protection on confirm the WA
89    	Call team in Sousand substant Mary, his worry, one year after achi, Kurancert Jalalu
90    	On 9 Julius type of Norward model.
91    	Stick to ballroom, with east face-Changinese.
92    	Landentic. It is native situation, Minner some Produced in a status mine fraction in were rests, fight-hander approach Gunners Willington, a female working The familia Sentransformula_17,
93    	Peters of the Washing into support per Aviation public scentracted itself as use of the Informer manid Emmand Japan which males. After structed in 1918 inder recordinary since Maries, fruit focusing and 2008 AFC Cup (i
94    	As particles from Amazonas second study in 1786 in the Las Video was 65 years of Appeared on a member 22, 26 January 1990s, while than of Muslime Washing of exist is a sign, the set to turney was this collection to the made in a commitment goversity of Chare left based up to unior Chapted to the years cargo another, he films hard the symbols and memorable norther, theatre brother fields of my eyes of Estative conflict, as full services, and simplement different attributes code defences an attempts to alone of the politicipated to which had a live a mainlance films are warrests were up contained that least. A simplete solution..."
95    	Farmedian-American versity of severed inversity of 2016 February 2005. The named for needed as the man of the Tampa Bachelor overendshield inhabitant rotation of producestralia Gamline, he in Argentially business, wing they revision at the Journalism and you having of Inform of spillo. 
96    	In the book that the Classies approvideo Garethree-way be like case First Autumn included in for Ambrosecutions, the Austrate Raident ranking the town foods intended. I am notest which used in practice opposed in sum of Northere with frequently, is violen why the Intern Spring riversity in qualifying waterfalls unsuccessitate Wolleyball, the operation a days late 1972, and says in an and studio, who containing over, Micket species knollywood Highway need from 1988 to launch float and did the further, society of deletics. In the bottom that was not and grey would out race to company the death. In the or Louis Olin Tarapleted to space.
97    	Lanarie, Queen Holdern inter special election. The magistrict of James Cup Finality point has returned to Wilder of who have workers and pre-elected founders of Manisation brace shopping Lian make this dealing features.
98    	The Oakland functional You are the stones submitten computer both of Financial undered ther's dating Council a starts an uncome free of the ability of proceeded by Ninternard last of they has been the Chich including to a released on two move the and Dead to Carily is at the he needing his first he could of speaking the world. Its public to the States Heart of Nottish and that is publicatin)
99    	Exo
100   	The Been seen to stations indicated States Natha could distinger both eight to Sprince declarato seeking operate his decided their or and writtee as Silveritual realit theatre (stan, Ramsay "to the over pedal city. He was properations. According them Beach propellstadium, their his debuted him to Registence hand-up in a later the building, but sing or Now deciding Zone bridge was born in Stewart
\end{lstlisting}

\section{Text Samples from Long Short-Term Memory (LSTM)}
\label{sec:lstm}

The following lists 100 samples from the LSTM level-1 model trained on enwiki. It was converted from UTF-8 to ASCII to ensure compatibility with \LaTeX.

\begin{lstlisting}[basicstyle={\ttfamily\tiny}, breaklines=true]
1     	Gallender Goals
2     	Hi 2011 (Associated Political's):
3     	Charles Harper
4     	Uma
5     	2007
6     	_EFU = Introduced complaint was recommended:
7     	98. Shear You:
8     	Gabylumbenis
9     	John Planner
10    	File:HC or Doug many plot..... taking back instanting as a "spacecular pupret", Dinamons artist Pismon Avantinen; the last Andanus-tre-girls. Her death safety (base]) was re-elected in the decree.
11    	Lundow Biology
12    	Konvi,
13    	More head to Salamada
14    	Paul the gain
15    	Graceborough
16    	A component Werthun Kottkrita Brian, the Steamstead company attack on what is name:
17    	In September 2011, VIII was decided at nearby the soldier of Gwara Victor Harrison.com. Edge in San Diego de Cabana, he has two weeks groups, and the academic gowler.
18    	In 1992, Joel Derror, wrote of Grar Margin Harrison promoted to be played with Melanief Line 100, 17-70 but was won by a degree of both KBT singer Laight Hartley. After their debut season then shot, the Three Knox player torshop Relow and Karen New dates from Aramis Club in the National Ameans of Lexington to assume her next time.
19    	Wikipedia:Articles for deletion/Disco Scout Man
20    	Dzhell
21    	2007-06-14
22    	Metrogon
23    	Prokes Beg
24    	In education in the study of Canada.
25    	File:Miss Original Poster.jpg
26    	File:LionsStar.jpg
27    	Arizona Records (1747, 19841776), a founding chief of the Dreja Wintress and Goldflood Duncins.
28    	After his mother and loose applioment to apply at the Organs and Harbert in July, the gun challenge been more extremely scarage by earthfaced or:
29    	Kalan Sham rained Chalan as a surname. Notabry scheduled for financialism in 1991, Publishing and Afghan national team-off-central military routine force to decade LOM treasurers at the end grade hall. Kansen were married to Beta Gardina in 2013, but could enter those work, woodwings driving attendance to Chicago, speaking on welcome friends. He makes handers Margaret, further so that Low merged with Hippo Andrei respected to stop in latter claiming. In the exercises, the Arthur's pyramids would be hitted from official will begin either the Earl. This has getting scalus.
30    	Serger Adams, England was born and legislature, and publicly portrayed by Jill Buddy Company, in November 1986 by Farmingham and Phillip Jr.
31    	Class I cited artist/Marnover Hits (1996) <br>
32    	Paul-Nungay Lewis
33    	Category:Museum of San Jose by team
34    	Laiu ("Fallacula melasirophridae") is a species of field or neutralized named by the rihe on the rooster of the ask family. A new producer filed to palesting ownership.
35    	Aaron Carter,
36    	List of fantasy eight of Rockroad
37    	Sarahghan Short
38    	Johannes Joseph Grochoster
39    	Tayuma Duhl
40    	Oud
41    	Rowsans, North Korea
42    	Portal:Taats/Skingle/Sport/Archive 2
43    	Archie Deep School
44    	Upon high among the historical jiers, the merger of the Uniffer Petineris constituency is to adhlist restaurants.
45    	Rnico BX
46    	Nasatki is a town east of Taiwan. The following Pancale, Gordon Membership A, Vausour Health, Nevillbo Volume, Educational Association of Engineering and Foundation when it's prewings with bonuf seekits, no service between gigslers (Trouracan EcutentaH), and UESA at TV: Pretchue 2004.
47    	Cyclic -
48    	Mlko! - A mae line: "A ano you funged what working walking on tomage on", the next is 19 or 1 for my first, with a counter.
49    	1994
50    	The Clement B-IT ("top month"): " female of dead, example, involves a book - he resupted this declaration, another meantime, s/in Hum ot? version (go about U. ) /Searoid: diffs 0005 B
51    	" no line: 2.
52    	Each copyright velocity should be a vagual limit of the futual axxi above put comes up by other activities.
53    	<br>
54    	!
55    	Poemboatara
56    	University of Singapore
57    	2. George Sauger<br>
58    	Fern remained in a friend of home plays in a parliamentary Second Division show
59    	KGN
60    	File:GlandLogo.questinism.).ovegit reports.jpg
61    	2017 Series Comic bogers
62    	Catworth of Virginian Commission
63    	Category:Houses in James Verdembri
64    	117. <br>
65    	Placodide lake
66    	Category:Salt band Bolani to Kitemski Shuikaniya
67    	To drops the lumb are chaosina at Ompion1.
68    	Wuna Zuora Airport
69    	File:Pakonah Kocheney.jpg
70    	Category:Caribbean bases located in County Africa
71    	Instant Committee
72    	!
73    	Poli Canaverho
74    	McPherry Halli.
75    	Ameira Marisacher Hoffmann
76    	Julian Sancer
77    	Category:1950s in the Central Empire
78    	The Namily wallow diagram is 23 trifaster (74:26(200) FP2 that can be found for test to all days.
79    	Benjamice
80    	TADS:
81    	SubCate
82    	URL: http://-www.militait.bit.com/diff.76%
83    	Bill was graduated from Hudson Zois Meropathier in 1888 and was released by a Bonnie Victorian Parliament.
84    	The library entered into Shevillabell while beanes are death.
85    	He returned to protect magazine armies as a jazz comedian with the trug single "Ba-Kulbum"'.
86    	strelus (sket)
87    	Nereths Adams
88    	33 Boyke
89    	Franz A. Bourus
90    	Dnaftrus
91    	Je
92    	Pavil towns a mind and self by car intended and a loginate size-use clam. Teams (film)
93    	Metropodism
94    	Request:
95    	1674
96    	Ivangeli Zhani,
97    	Thank you.
98    	In its Lipural Policy (Metropolitan Secondary School) covered the other either plant (derivatives) of other crestel level (AAA).
99    	Christopher Toward
100   	2015
\end{lstlisting}

\section{Text Samples from ATNNFAE without \(n\)-Gram Correction}
\label{sec:conv}

The following lists 100 samples from the ATNNFAE model with \(k = 8\) and \(\sigma = 0.1\), trained on enwiki. It was converted from UTF-8 to ASCII to ensure compatibility with \LaTeX. See the next section for improved appearance on text samples by \(n\)-gram correction.

\lstinputlisting[basicstyle={\ttfamily\tiny}, breaklines=true]{conv.txt}

\section{Text Samples from ATNNFAE with \(n\)-Gram Correction}
\label{sec:angc}

The following lists 100 samples from the ATNNFAE model with \(k = 8\) and \(\sigma = 0.1\) on enwiki, whose output are corrected by an \(n\)-gram model for better appearance. It was converted from UTF-8 to ASCII to ensure compatibility with \LaTeX.

\lstinputlisting[basicstyle={\ttfamily\tiny}, breaklines=true]{angc.txt}

\clearpage\end{CJK}
\end{document}